\documentclass{article}

\usepackage[final]{corl_2022} %
\usepackage{amsmath}
\usepackage{amsfonts}
\usepackage{amssymb}
\usepackage[font=small]{caption}
\usepackage{graphicx}
\usepackage{subcaption}

\renewcommand{\vec}{\bm}
\newcommand{\tran}{^{T}}
\usepackage{bm}

\title{Learning Semantics-Aware Locomotion Skills\\ from Human Demonstration}

\author{
  Yuxiang Yang$^{12}$\thanks{Work done during internship at Google.}, Xiangyun Meng$^1$, Wenhao Yu$^2$, Tingnan Zhang$^2$, Jie Tan$^2$, Byron Boots$^1$\\
  $^1$University of Washington, $^2$Robotics at Google\\
  \texttt{\{yuxiangy,xiangyun,bboots\}@cs.washington.edu}\\
  \texttt{\{magicmelon,tingnan,jietan\}@google.com}
}

\begin{document}
\maketitle

\begin{abstract}
The semantics of the environment, such as the terrain types and properties, reveal important information for legged robots to adjust their behaviors.
In this work, we present a framework that uses semantic information from RGB images to adjust the speeds and gaits for quadrupedal robots, such that the robot can traverse through complex offroad terrains.
Due to the lack of high-fidelity offroad simulation, our framework needs to be trained directly in the real world, which brings unique challenges in sample efficiency and safety.
To ensure sample efficiency, we pre-train the perception model on an off-road driving dataset.
To avoid the risks of real-world policy exploration, we leverage human demonstration to train a speed policy that selects a desired forward speed from camera image.
For maximum traversability, we pair the speed policy with a gait selector, which selects a robust locomotion gait for each forward speed.
Using only 40 minutes of human demonstration data, our framework learns to adjust the speed and gait of the robot based on perceived terrain semantics, and enables the robot to walk over 6km safely and efficiently.
\end{abstract}

\keywords{Legged Locomotion, Semantic Perception, Imitation Learning, Hierarchical Control} 

\section{Introduction}

In order to operate in complex offroad environments, it is crucial for quadrupedal robots to adapt their motion based on the perception of the terrain ahead.
When encountering new terrains, the robot needs to identify changes in key terrain properties, such as friction and deformability, and respond with the appropriate locomotion strategy to maintain a reasonable forward speed without incurring failures.
In many cases, information about such terrain properties is more easily inferred from a terrain's \emph{semantic} class (e.g. grass, mud, asphalt, etc.), instead of its \emph{geometric} shape (e.g. slope angle, smoothness) \cite{affordance_theory, semantic_affordance_cvpr}.
However, recent works in perceptive locomotion \cite{googlevisuallocomotion, gangapurwala2020rloc, fankhauser2018robust, villarreal2020mpc, jenelten2020perceptive, eth_hike, rma_navigation} mostly focus on the \emph{geometric} aspect of the terrain, and do not make use of such \emph{semantic} information.

In this work, we present a framework for quadrupedal robots to adapt locomotion behaviors based on perceived terrain semantics.
The central challenge in learning such a semantic-aware locomotion controller is the high cost in data collection.
On one hand, while simulation has become an effective data source for many robot learning tasks, it is still difficult to build simulation for offroad environments, because modeling the complex contact dynamics accurately and rendering the environment photorealistically in such offroad terrains is not yet possible in simulation.
On the other hand, data collection in the real world is time-consuming and requires significant human labor.
Moreover, the robot needs to remain safe during the data collection process, as any robot failure can cause significant damage to the hardware and surrounding environment.
Therefore, it is difficult to use standard reinforcement learning methods for this task.

Our framework addresses all concerns above, and learns semantics-aware locomotion skills directly in the real world.
To reduce the amount of data required, we pre-train a semantic segmentation network on an off-road driving dataset, and extract a semantic embedding from the model for further fine-tuning.
To avoid policy exploration in real-world environments, we collect speed choices from human demonstrations, and train the policy using imitation learning \cite{pomerleau1988alvinn}.
Additionally, inspired by previous results on the relationship between speed and gait in animals \cite{hoyt1981gait} and legged robots \cite{fast_and_efficient, fu2021minimizing}, we pair the speed policy with a gait selector to further improve the robot's stability.
With the pre-trained image embedding, the imitation learning setup, and the gait selector, our framework learns semantics-aware locomotion skills directly in the real world safely and efficiently.

We deploy our framework on an A1 quadrupedal robot from Unitree \cite{a1robot}. Using only 40 minutes of human demonstration data, our framework learns semantics-aware locomotion skills that can be directly deployed for offroad operation.
The learned skill policy inspects the environment and selects a fast and robust locomotion skill for each terrain, from slow and cautious stepping on heavy pebbles to fast and active running on flat asphalts.
The learned framework generalizes well, and operates without failure on a number of trails not seen during training (over 6km in total).
Moreover, our framework outperforms manufacturer's default controller in terms of speed and safety.
We further conduct ablation studies to justify the important design choices.

The technical contributions of this paper include:
\begin{enumerate}
    \item We develop a hierarchical framework that adapts locomotion skills from terrain semantics.
    \item We propose a safe and data-efficient method to train our framework directly in the real world, which only requires 40 minutes of human demonstration data.
    \item We evaluate the trained framework on multiple trails spanning 6km with different terrain types, where the robot reached high speed and walked without failures.
\end{enumerate}

\section{Related Works}

\paragraph{Perception for Legged Robots}
Creating a perceptive locomotion controller is a critical step to enable legged robots to walk in offroad, unstructured environments. 
Most importantly it allows robots to detect and react to terrain changes proactively before contact.
Many prior works have focused on understanding terrain \emph{geometry} from perceptive sensors \cite{fankhauser2018probabilistic, eth_traversability, googlevisuallocomotion, jenelten2020perceptive, villarreal2020mpc}. 
However, such information can be insufficient as it does not reveal important terrain properties such as deformability or contact friction \cite{eth_hike, rma_navigation, transformer_a1, wellhausen2019should}.
To ameliorate this, recent works proposed to update this geometric understanding of a terrain with proprioceptive information \cite{eth_hike, rma_navigation, transformer_a1}.
However, these methods sacrifice proactivity, as the update cannot happen until \emph{after} the robot has stepped on the terrain.

Another approach is to infer the terrain properties from its \emph{semantics}~\cite{lalonde2006natural, khan2011high, schilling2017geometric, shaban2022semantic}, so that the robot can detect changes in terrain property \emph{before} contact, and select its locomotion strategies proactively.
Recently, \citet{terrain_segmentation} trained models to predict terrain class and roughness, and used the prediction to modulate the height and navigational path of a wheel-leg hybrid robot in an indoor environment.
Our framework uses a similar semantics-based approach in the perception module, and extends the result to off-road environments with a wide variety of terrains by adapting both the speed and gait of the robot.

\paragraph{Terrain Traversability Estimation} 

The goal of our perception module is to assess the traversability of the terrain ahead of the robot.
Researchers have proposed a number of approaches to estimate traversability from perception data, including manually designed \cite{fan2021step}, learned from self exploration \cite{eth_traversability, terrain_classification}, or learned from human demonstration \cite{jphow}.
While learning-based approaches provide more flexibility, they usually require large amounts of data, which is difficult to collect in the real world. As a result, most approaches rely on simulation \cite{dosovitskiy2017carla} as a source for training data.
However, simulation is not feasible for our task, as it is currently difficult to accurately model the complex contact dynamics and create photorealistic renderings of off-road environments.
Unlike previous approaches, our framework can be trained directly in the real world, and requires only 40 minutes of human demonstration data.

\paragraph{Motion Controller Design for Perceptive Locomotion}
Another important question in perceptive locomotion is the design of a motion controller that effectively makes use of the perceptive information.
A common strategy is to create a low-level motion controller that plans precise foothold placements based on the perceived terrain \cite{googlevisuallocomotion, gangapurwala2020rloc, fankhauser2018robust, villarreal2020mpc, jenelten2020perceptive}. While these methods have shown good results in highly uneven terrains, the high computational cost required for terrain understanding and rapid planning makes it infeasible for complex offroad environments.
In this work, we devise a novel way to interface between perception and low-level motion controllers for legged robots, where the high-level perception model outputs the desired locomotion skills, including forward speed and robot gait, to a low-level motor controller.
With our framework, the robot can select a safe and fast walking strategy for different terrains, which is crucial for offroad traversal.

\section{Overview}
\begin{figure}[t]
    \centering
    \vspace{-1em}
    \includegraphics[width=0.7\linewidth]{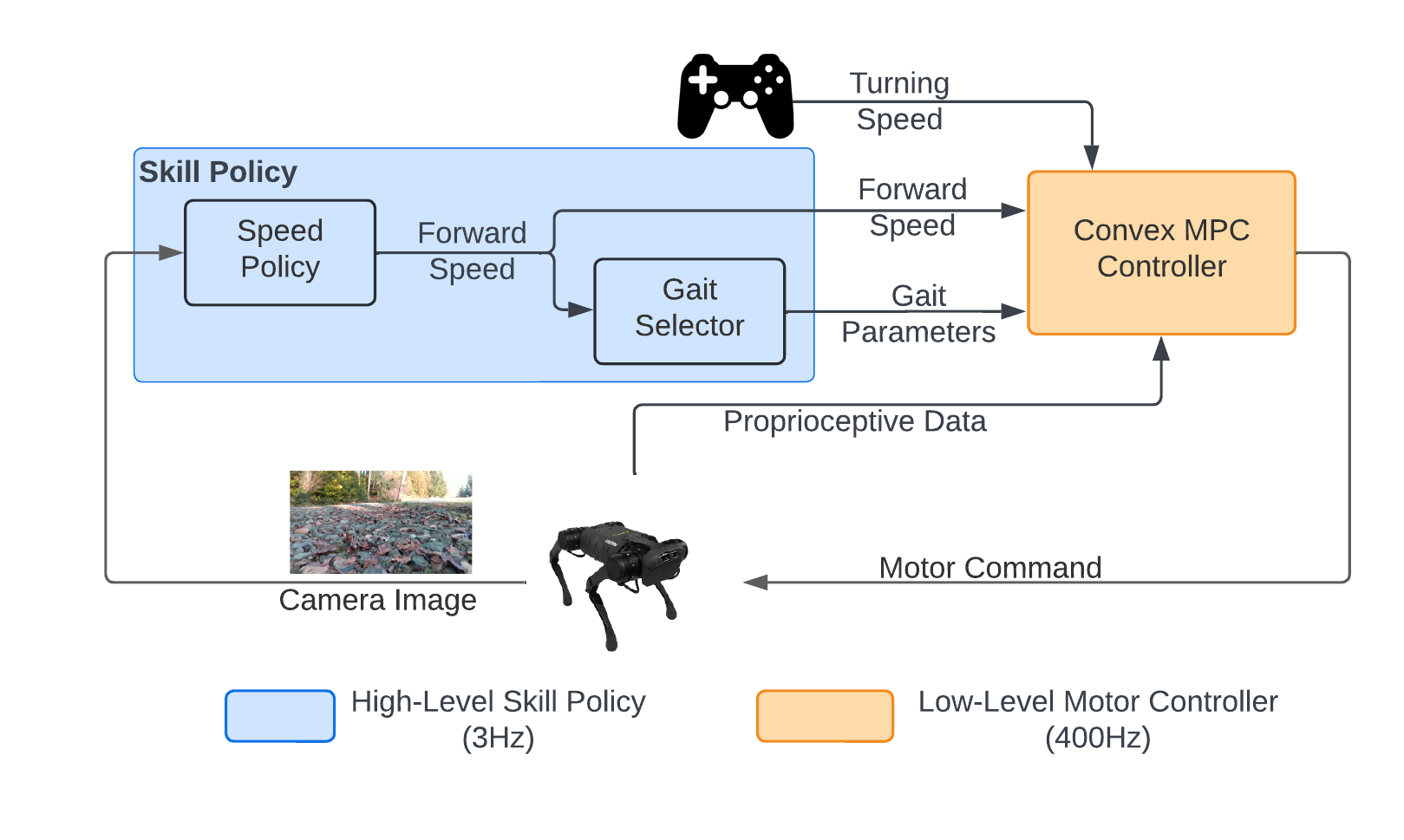}
    \vspace{-1em}
    \caption{Our framework consists of a high-level skill policy and a low-level motor controller. The skill policy selects locomotion skills (gait and speed) based on camera images. The low-level controller computes motor commands for robot control.}
    \vspace{-1em}
    \label{fig:block_diagram}
\end{figure}
Our hierarchical framework (Fig.~\ref{fig:block_diagram}) consists of a high-level skill policy and a low-level motor controller. At the high level, the skill policy receives the RGB image stream from onboard camera and determines the corresponding locomotion skill. Each skill consists of a desired forward speed and a corresponding locomotion gait, which are computed by the speed policy and gait selector, respectively. 
We train the speed policy using imitation learning from human demonstrations, and manually design the gait selector to find the appropriate gait for each forward speed. At the low level, a convex MPC controller \cite{mitcheetahmpc} receives the skill command from the skill policy, and computes motor commands for robot control. In addition, the convex MPC controller can optionally receive a steering command from an external teleoperator, which specifies the desired turning rate.

\section{Learning Speed Policies}
\label{sec:speed_policy}
In unstructured offroad terrains, it is important for a robot to adjust its speed in response to terrain changes, so that it can traverse through different terrains efficiently and without failure.
To achieve that, we design a speed policy, which computes the desired forward speed of the robot based on camera images. 
We train the speed policy using a two-staged procedure:
First, we pre-train a semantic embedding from an offline dataset.
After that, we collect human demonstrations and train the speed policy using imitation learning.

\subsection{Pre-trained Semantic Embedding}
\begin{figure}[t]
    \centering
    \vspace{-1em}
    \includegraphics[width=0.85\linewidth]{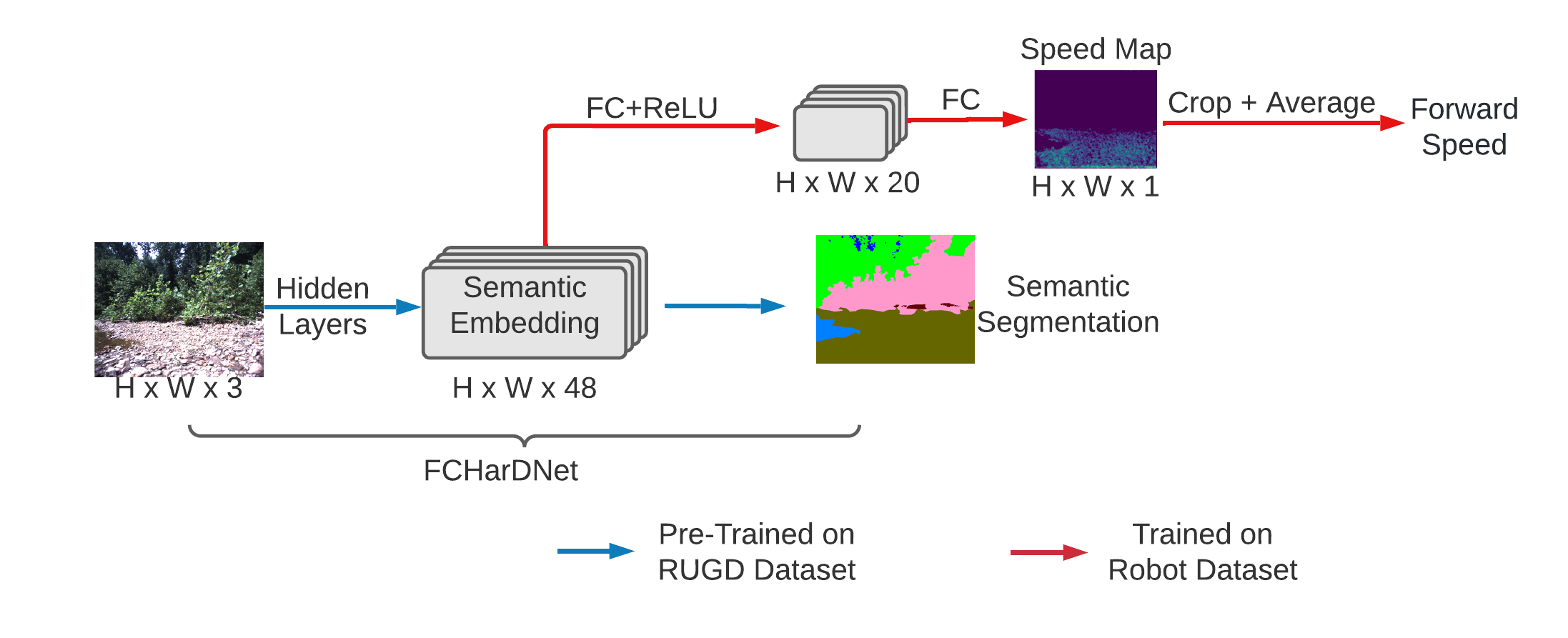}
    \vspace{-1em}
    \caption{Architecture of our perception model. We extract a semantic embedding from a pre-trained semantic segmentation network, and use it to learn and predict forward speeds.}
    \vspace{-1em}
    \label{fig:perception_architecture}
\end{figure}
\label{sec:semantic_embedding}
To reduce the amount of real-world data required to train the speed policy, we pre-train a semantic segmentation model and extract a semantic embedding for subsequent finetuning.
We implement the model based on FCHarDNet-70~\cite{fchardnet}, which is a compact fully-convolutional encoder-decoder architecture with good real-time performance.
We pre-train the model on the RUGD dataset \cite{RUGD}, an off-road driving dataset with pixel-wise semantic labels (grass, dirt, rock, etc.).
We choose RUGD because of its similarity to the images collected by the robot camera.

The next step is to extract an embedding from the pre-trained FCHarDNet~\cite{fchardnet} model for finetuning on robot data. 
Although the pre-trained model achieves good performance on the RUGD dataset, its predicted segmentation becomes less accurate on robot images due to distribution shift. 
Meanwhile, the output of the hidden layers still provides a continuous semantic description for each pixel.
Therefore, we extract a \emph{semantic embedding} from the output of the last hidden layer in FCHarDNet, which assigns a 48-dimensional embedding vector to each pixel in the input image (Fig.~\ref{fig:perception_architecture}). 
We then compute a speed map by feeding the embedding of each pixel through a fully-connected layer, and compute the forward speed by averaging over a fixed region at the bottom of the speed map, which roughly corresponds to a rectangular area 1m long, 0.3m wide in front of the robot.
The speedmap provides a straightforward and intuitive way to understand the model's predictions, and can be used in navigational tasks such as path planning.

\subsection{Learning Speed Commands from Human Demonstration}
Even with the pre-trained semantic embedding, finding the appropriate speeds for offroad terrains using reinforcement learning is still challenging, due to omnipresent noise and safety concerns in the real world. 
As an alternative, a human operator can readily assess the robot's stability and adjusts the speed command accordingly, based on the operator's previous experience with the robot platform.
Therefore, we collect speed commands from human demonstrations and train the speed policy using imitation learning.

We collect human demonstrations by tele-operating the robot on a variety of terrains, including asphalt, pebble, grass and dirt.
During data collection, the operator gives speed command using a joystick, while other components of the pipeline, such as the gait selector and motor controller, functions accordingly (Fig.~\ref{fig:block_diagram}).
Each time the camera captures a new image, we store the image and the corresponding speed command. 
We then train the speed policy using behavior cloning \cite{pomerleau1988alvinn}, where the objective is to minimize the difference between predicted speed and human command. 

\section{Speed-based Gait Selector}
In addition to speed, the \emph{gait} of a legged robot, such as its foot swing height, can greatly affect its traversability, especially on uneven terrains.
While the perception policy can output speed and gait parameters jointly, training such a policy using imitation learning can be challenging, as it is difficult for the human operator to demonstrate speed and gait choices at the same time.
Meanwhile, previous studies in animal \cite{hoyt1981gait} and robot \cite{fast_and_efficient, fu2021minimizing} locomotion have revealed a close connection between speed and gait choices.
Inspired by this discovery, we simplify the demonstration and learning process by designing a heuristic-based \emph{gait selector}, which computes the appropriate gait parameters based on desired forward speed.

\paragraph{Gait Parameterization}
In our design, each gait is parameterized by three parameters, stepping frequency (SF), swing foot height (SH), and base height (BH). The \textbf{stepping frequency (SF)} determines the number of locomotion cycles each second. 
Similar to~\cite{fast_and_efficient}, we adopt a phase-based parameterization for gait cycles, where each leg alternates between swing and stance.
In addition, we assume a trotting pattern for leg coordination, where diagonal legs move together and are $180^{\circ}$ out-of-phase with the other diagonal.
The trotting pattern is known for its stability and thereby is the default gait choice in most quadrupedal robots \cite{mitcheetahmpc, a1robot}. The \textbf{swing foot height (SH)} determines the leg's maximum ground clearance in each swing phase. While a higher swing height improves stability on uneven terrains by preventing unexpected contacts, a lower swing height is usually necessary for high-speed running. The \textbf{base height (BH)} specifies the height of robot's center-of-mass. While a low base height gives better stability at high speeds, a higher base height can be beneficial when traversing through unknown obstacles. 
\vspace{-2mm}
\paragraph{Speed-Based Gait Selection}
\begin{figure}[t]
    \centering
    \vspace{-1em}
    \includegraphics[width=0.8\linewidth]{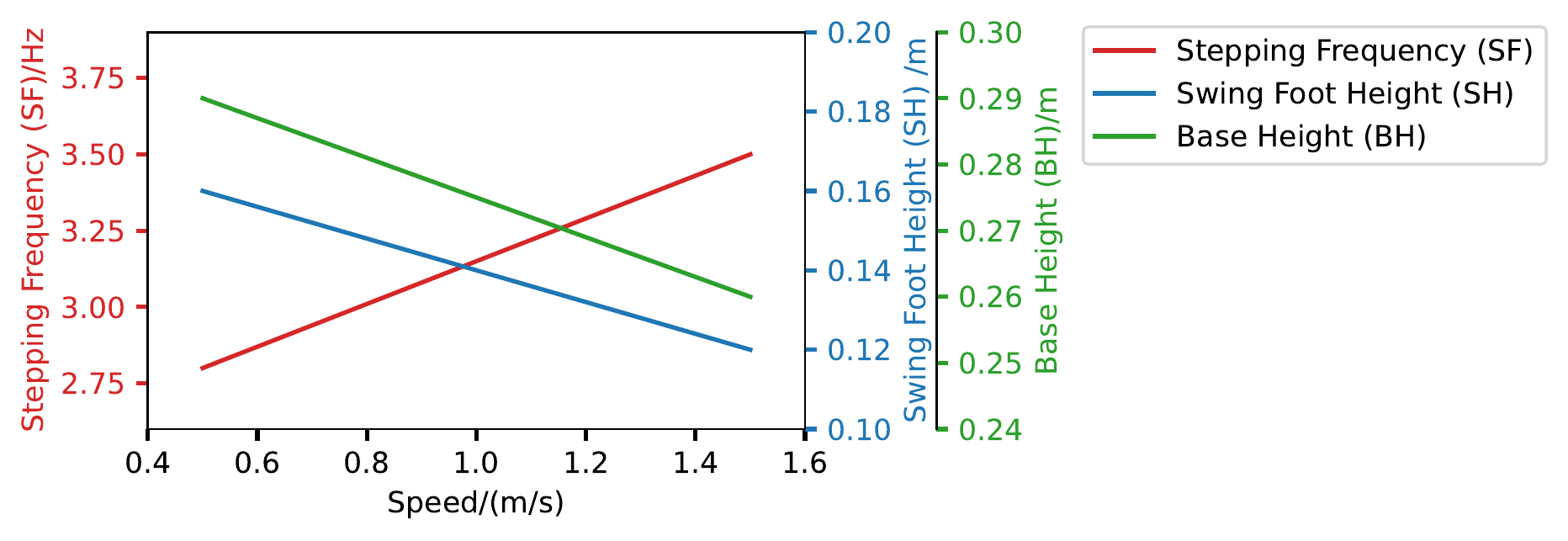}
    \caption{The gait selector selects gait parameters (SF, SH and BH) based on desired forward speed. For example, when the desired speed is 0.5m/s, the speed selector would choose a stepping frequency of 2.8Hz, a swing foot height of 0.16m, and a base height of 0.29m.}
    \label{fig:speed_selector}
    \vspace{-1em}
\end{figure}
\label{sec:gait_selector}
We use empirical evidence to design the speed-based gait selector, which finds a gait with high traversability for each speed.
More specifically, for the boundary speeds (0.5m/s and 1.5m/s), we first try different SFs with a nominal SH (0.12m) and BH (0.26m) , and find the lowest SF that would still ensure base stability (2.8Hz and 3.5Hz).
After that, we sweep over different values of SH and BH, to find the highest value of both that would allow the robot to walk robustly without falling.
Lastly, we linearly interpolate the parameter values between the boundary speeds to find the gait for intermediate speeds.
See Fig.~\ref{fig:speed_selector} for details.

\section{Low-level Convex MPC Controller}
The low-level convex MPC controller computes and applies torques to each actuated degree of freedom, given the locomotion skills from the skill policy. 
Our low-level convex MPC controller is based on \citet{mitcheetahmpc} with two important modifications.
Firstly, due to the robot's small form factor, it needs to constantly re-orient its body on uneven terrains, such as bumps and potholes.
Therefore, we implemented a state estimator to estimate ground orientation, and adjust the robot pose to fit the ground, similar to \citet{ethslopes}.
Secondly, to reduce foot slipping, we implement an impedance controller for stance legs \cite{ethimpedance}.
In addition to the motor torque command computed by MPC, the impedance controller adds a small feedback torque to track the leg in its desired position.
We found both techniques to significantly improve locomotion quality.
Please refer to Appendix \ref{sec:appendix_mpc} for details.

\section{Experiment and Result}

To see whether our framework can learn to adapt locomotion skills based on terrain semantics, we deploy it to a quadruped robot and test it in a number of outdoor environments in the real world. We aim to answer the following questions in our experiments:
\begin{enumerate}
    \item Can our framework operate without failure in complex offroad terrains for extended periods of time, and how does it compare with existing baselines?
    \item How does our framework generalize to terrain instances not seen during training?
    \item Can our framework walk at high speed while ensuring safety?
    \item What are the important design choices in training the perception module and how does it affect performance?
\end{enumerate}

\subsection{Experiment Setup}
We implement our framework on an A1 quadrupedal robot from Unitree \cite{a1robot}. 
We equip the robot with an Intel Realsense D435i camera to capture RGB images, and a GPS receiver to track its real-time location.
We implement the entire control stack in the Robot Operating System (ROS) framework \cite{ROS}, and deploy it on a Mac Mini with M1 chip, which is mounted on the robot. The convex MPC controller runs at 400Hz, and the speed policy and gait selector run at 3Hz.

To train the speed policy, we collected 7239 frames of data on a variety of terrains, which corresponds to 40 minutes of robot operation. The entire process, including robot setup, data collection and battery swaps, took less than an hour. The speed policy is trained on a standard desktop computer with an Nvidia 2080Ti GPU, which took approximately 20 minutes to complete.

\subsection{Fast and Failure-free Walking on Multiple Terrains}
\begin{figure}
    \centering
    \vspace{-1em}
    \includegraphics[width=.8\linewidth]{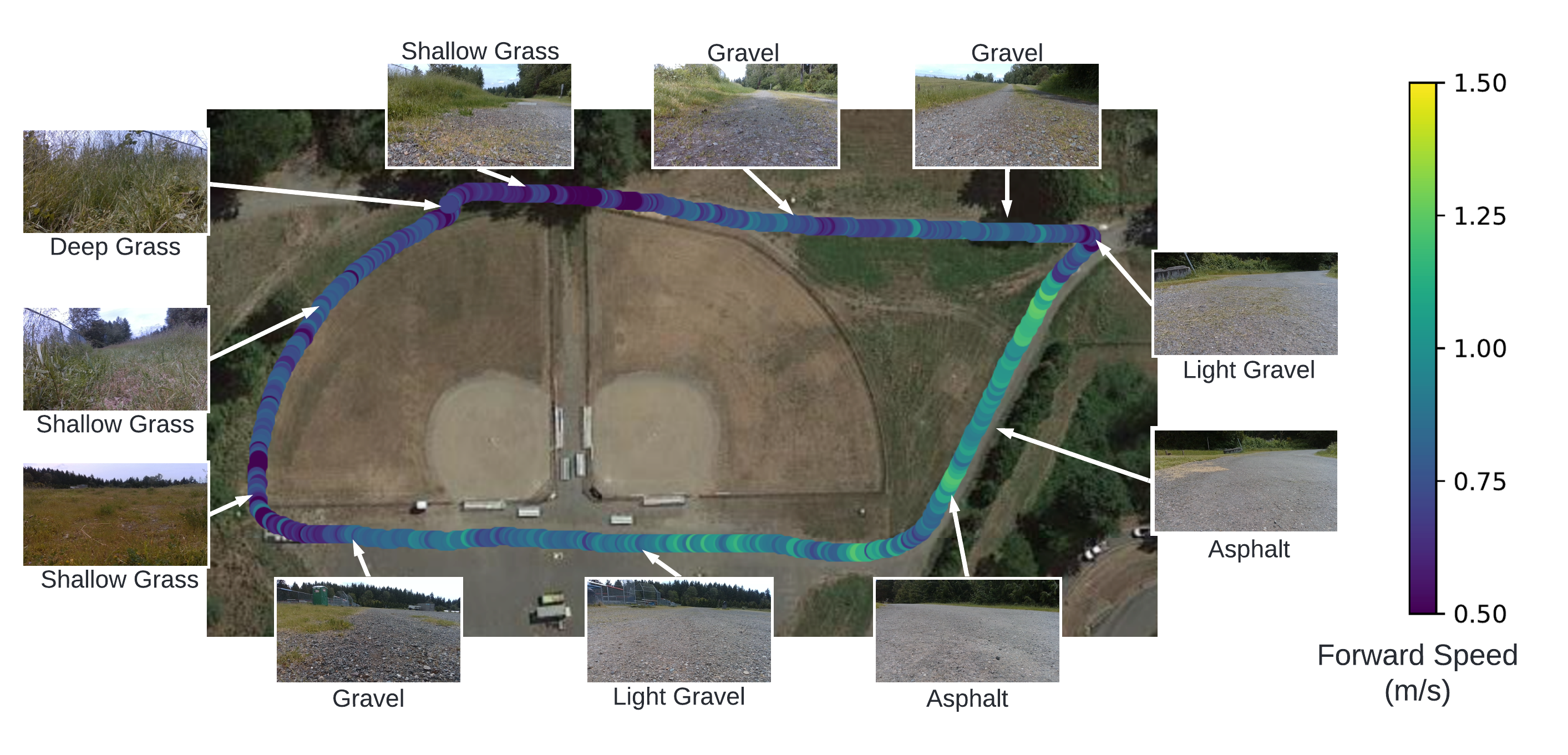}
    \label{fig:ablation_overview}
    \vspace{-1em}
    \caption{    \label{fig:ablation_overview}
The 450m-long test trail consists of multiple terrain types such as deep grass, shallow grass, gravel and asphalt. The learned skill policy adjusts the speed and gait based on terrain semantics, and walks faster on easier terrains.}
    \vspace{-1em}
\end{figure}

To evaluate the adaptivity of our framework, we test our framework on an outdoor trail with multiple terrain types, including deep grass, shallow grass, gravel and asphalt (Fig.~\ref{fig:ablation_overview}). 
Our controller switches between a wide range of skills as it traverses through the trail, from slow and careful stepping to fast and active walking, and completes the 450m-long trail in 9.6 minutes, comparable to the performance of human demonstration (10 minutes).

\begin{table}[t]
    \centering
    \small
    \begin{tabular}{c|c|c|c|c}
         \hline
         \textbf{Policy Type}&\textbf{Speed} & \textbf{Gait Params} & \textbf{Traversal Time} & \textbf{Number of Failures}\\
         & (m/s) & (SF, SH, BH) & (min) & \\
         \hline
         Fixed-Slow&0.5 & [2.8, 0.16, 0.29] & 15 & \textbf{0} \\
         Fixed-Mean&0.8&[3.0, 0.15, 0.28] & $\infty$ & 3\\
         Fixed-Medium&1& [3.1, 0.14, 0.28] & $\infty$ & 4 \\
         Fixed-Fast&1.5& [3.5, 0.12, 0.27] & $\infty$ & 10+ \\\hline
         Speed-Only&Adaptive & [3.1, 0.14, 0.28] &$\infty$&9\\
         Gait-Only&0.8 & Adaptive &$\infty$&2\\\hline
         Unitree-Normal &Tele-operated&N/A&11$\pm$0.4&\textbf{0}\\
         Unitree-Sport &Tele-operated&N/A&$\infty$&2\\\hline
         Fully-Adaptive (ours) &Adaptive & Adaptive & \textbf{9.6$\pm$0.2} & \textbf{0}  \\\hline
    \end{tabular}
    \vspace{1mm}
    \caption{Performance of different policies on the test trail (450m). Compared to other policies, our framework completes the entire trail without failure in the shortest time. We repeat the Unitree-Normal and Fully-Adaptive policies 3 times and report the mean and standard deviation of the traversal time. We do not repeat the other policies due to excessive robot damages.}
    \vspace{-1em}
    \label{tab:ablation_table}
\end{table}
We compared our learned framework with the following baselines on the same test trail (Fig.~\ref{fig:ablation_overview}), including Unitree's built-in controllers, and variants of our controller with no or limited adaptation. The result is summarized in Table.~\ref{tab:ablation_table}. Please refer to Appendix.~\ref{section:failure_locations} for further details.

\paragraph{Unitree's built-in controllers}
We tested two modes of the built-in controller, a \emph{normal} mode (Unitree-Normal) that walks up to 1m/s, and a \emph{sports} (Unitree-Sport) mode that walks up to 1.5m/s.
Both controllers do not include perception and assume a fixed gait at all times. 
Although normal mode completed the entire trail without failure, it walks slower than our learned framework, especially on asphalts, due to limitations on the maximum speed.
On the other hand, the sports mode controller failed to complete the course, and got stuck in deep grass twice due to insufficient swing foot clearance.

\paragraph{Fixed Skill with No Adaptation} 
For these baselines, we disable the perception module and operate the robot with a fixed locomotion skills.
We test four skills, namely slow, mean, medium and fast, operating at 0.5m/s, 0.8m/s, 1m/s, 1.5m/s, respectively, with the corresponding gait selected according to Fig.~\ref{fig:speed_selector}.
The slow, medium and fast skills cover the range of possible speeds achievable by our low-level controller, and the mean skill walks at a speed similar to the average speed achieved by our adaptive policy (0.78m/s).
The mean, medium skill and fast skill fail to complete the trail and incurred failures.
While the slow skill completes the trail without failure, its traversal time is 50\% longer compared to our learned framework.
\paragraph{Adapt Speed or Gait Only}
In our framework, we design a robot skill to be a combination of gait and forward speed.
To justify this design, we design two policies, where the robot adapts the gait or the forward speed only.
For the speed-only policy, we fix the gait parameters as if the forward speed is 1.0m/s in Fig.~\ref{fig:speed_selector}, and adapt the speed using our framework.
For the gait-only policy, we fix the base speed to be 0.8m/s, similar to the average speed attained by our learned policy, and adapt the gait using our framework.
Both policies failed to complete the trail.
For speed-only policy, we found the fixed gait to only work well when the base speed is close to 1m/s, and fails frequently at either higher or lower speeds.
For the gait-only policy, the robot manages to walk through most of the trail, but slips twice on rocky terrains.

\subsection{Generalization to Unseen Terrain Instances}

\begin{figure}
    \centering
    \includegraphics[width=.9\linewidth]{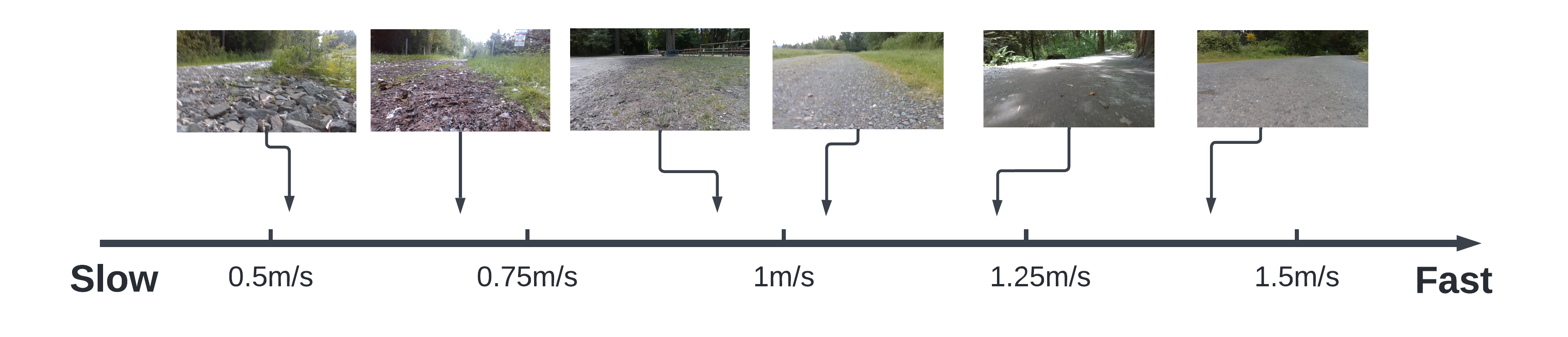}
    \vspace{-1em}
    \caption{Desired speed computed by the skill policy. The policy prefers faster skills for rigid and flat terrains, and prefers slower skills for deformable or uneven terrains.}
    \label{fig:speed_spectrum}
    \vspace{-1em}
\end{figure}
To further test the generalizability of our framework, we deploy the robot on a number of outdoor trails not seen during training.
The trails contain diverse terrain types, such as dirt, gravel, mud, grass and asphalt.
The robot traverses through these test trails without failure, and adjusts its locomotion skills based on terrain semantics.
Please refer to Appendix.~\ref{sec:appendix_generalization} for details.
To demonstrate the skill choices of our framework, we select a few key frames from the camera images and plot the corresponding speed in Fig.~\ref{fig:speed_spectrum}.
In general, the skill policy selects a faster skill on rigid and flat terrains, and a slower speed on deformable or uneven terrains.
At the time of writing, the robot has traversed through over 6km of outdoor trails without failure.

\subsection{Analysis on Speed and Safety}
\begin{figure}
    \centering
    \vspace{-1em}
    \begin{subfigure}{\linewidth}
        \centering
        \includegraphics[width=0.8\linewidth]{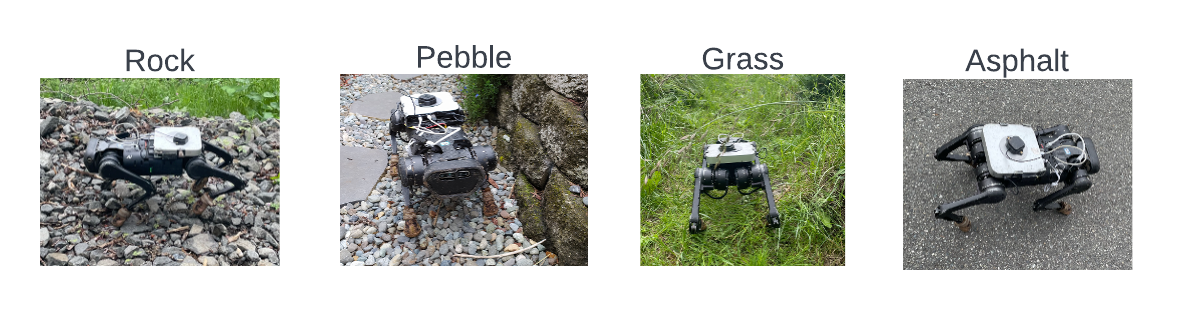}
        \vspace{-1em}
        \caption{We test our robot on different terrain types.}
        \label{fig:terrain_types}
    \end{subfigure}
    \begin{subfigure}{\linewidth}
        \centering
        \includegraphics[width=0.85\linewidth]{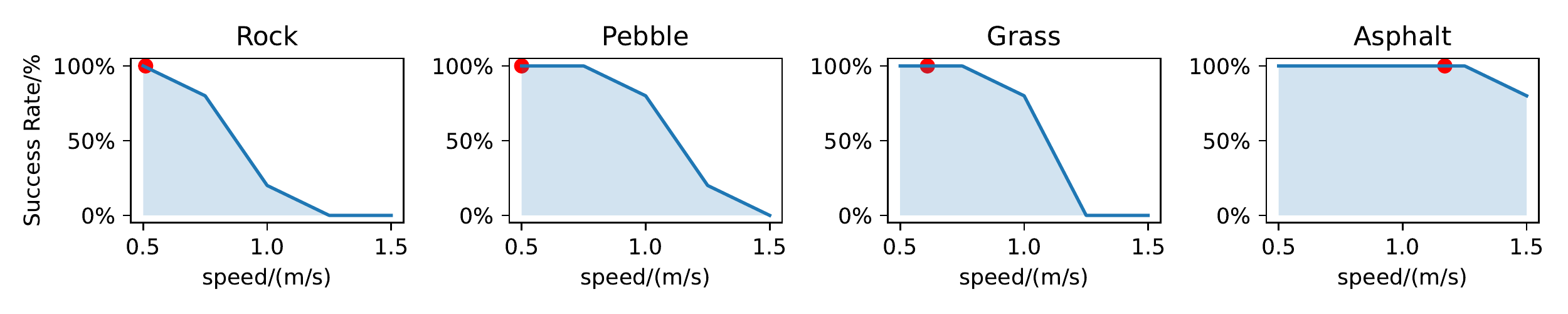}
        \vspace{-1em}
        \caption{Success rate of fixed locomotion skills (blue) and the learned policy (red).}
        \label{fig:terrain_success_rates}
    \end{subfigure}
    \caption{Our framework learns fast and safe locomotion skills. \emph{Top:} We deployed our skill policy to 4 different terrains. \emph{Bottom:} Our policy finds a high speed in the safe region of each terrain.}
    \label{fig:terrain_specific}
    \vspace{-1em}
\end{figure}
To test the performance of the learned skill policy in terms of speed and stability, we deploy the learned skill policy on four different terrains, including rock, pebble, grass and pebble (Fig.~\ref{fig:terrain_types}).
We compare our semantics-aware skill policy with 5 fixed skills, where the speed linearly interpolates between 0.5m/s and 1.5m/s. For each speed, the corresponding gait is selected according to Fig.~\ref{fig:speed_selector}.
For each terrain and skill combination, we repeat the experiment 5 times, and report the success rate, where a trial is considered successful if the robot does not fall over during the traversal (Fig.~\ref{fig:terrain_success_rates}).
By comparing the success rate at different speeds, we obtained an approximation of the safe speed range for each terrain.
We then test the performance of our framework by comparing the average speed obtained by our learned skill policy on each terrain against these safe speed ranges.

The maximum safe speed varies significantly on different terrains. 
For example, while the robot can walk up to 1.25m/s without failure on asphalt, it can only walk up to 0.5m/s on rock, due to unexpected bumps and foot slips on the surface. Although not directly optimized for speed or robot safety, our learned policy finds a close-to-maximum speed in the safe region of each terrain after learning from human demonstrations.
We also noted that on pebble and grass, there is a slightly larger gap between the maximum safety speed and the speed selected by the skill policy.
One reason for this is that the speed demonstrated by the human operator can be more conservative than the maximum safety speed.

\subsection{Ablation Study on Perception Module}
We compare our way of training the perception-enabled speed policy with a few baselines, which either trains the policy from scratch without pre-training, or extract the pre-trained embedding directly from the predicted semantic classes. We find that our policy, which is fine-tuned from output of hidden layer, achieves the smallest error on the validation set and predicts the speed map with high precision. Please see Appendix~\ref{sec:appendix_perception} for further details.

\section{Limitations and Future Work}
In this work, we present a hierarchical framework to learn semantic-aware locomotion gaits from human demonstrations.
Our framework learns to adapt locomotion skills for a variety of terrains using 40 minutes of human demonstration, and enables a robot to traverse over 6km of outdoor terrains without failure.
One limitation of our framework is that, while our robot walks robustly on a variety of off-road terrains, its performance is limited by the low dimensionality of human demonstrations.
For more difficult terrains such as steps or gaps, the robot will need more agile behaviors such as jumping, which requires a deeper integration between the perception system and low-level motor controller and learning more skills than speed or gait demonstrations.
Another limitation is that the perception system assumes that there is no non-traversable obstacles ahead of the robot, and therefore does not adjust the heading of the robot.
In future work, we plan to increase the agility of our controller and integrate path planning into our framework, so that the robot can operate fully autonomously in challenging off-road environments.

\clearpage
\bibliography{example}  %
\clearpage
\appendix
\section{Details of the Convex MPC Controller}
\begin{figure}[t]
    \centering
    \includegraphics[width=0.8\linewidth]{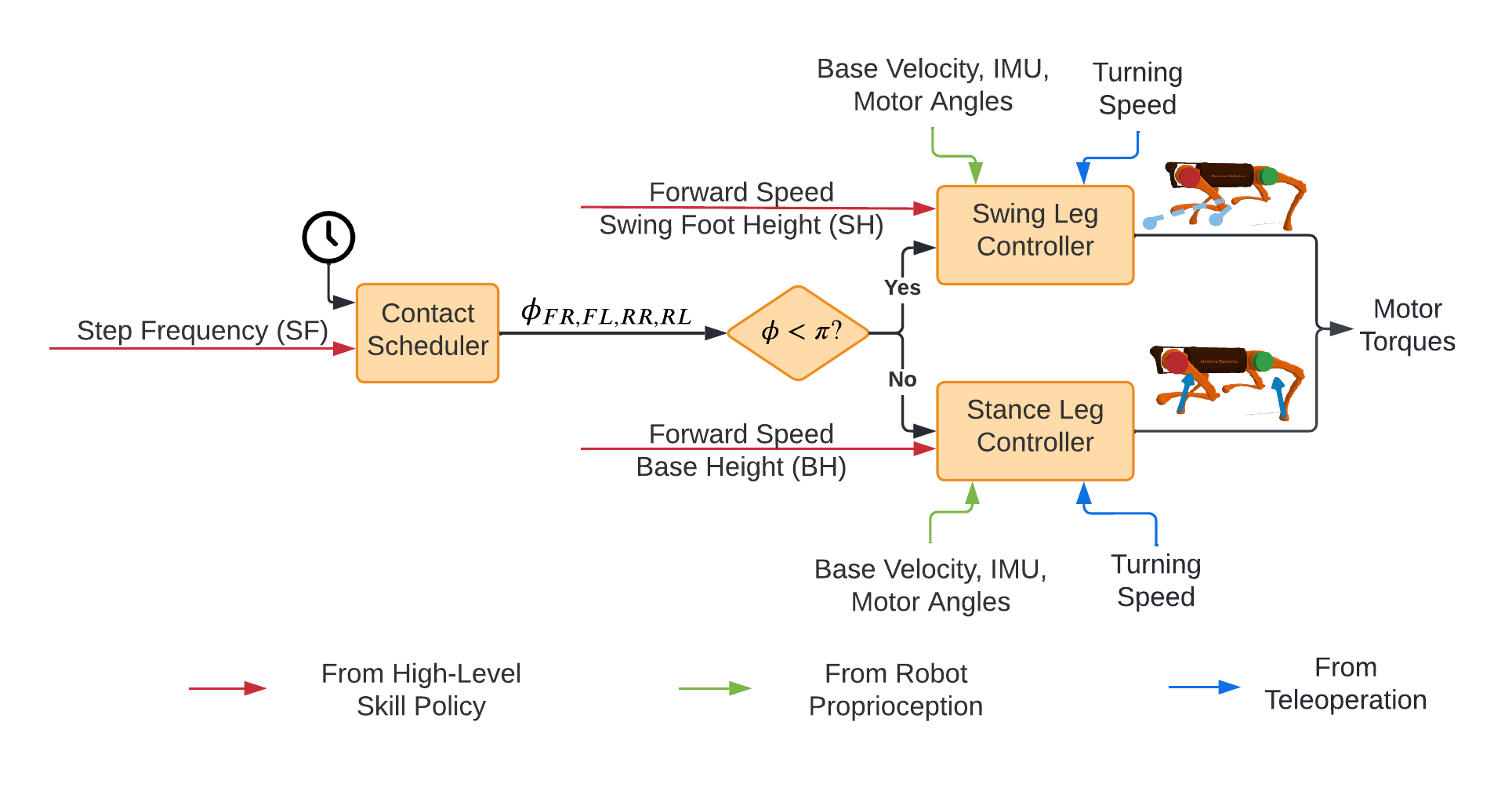}
    \caption{Overview of our low-level convex MPC controller. We use separate control strategies for swing and stance legs, and use a phase-based contact scheduler to determine the contact state of each leg.}
    \label{fig:my_label}
\end{figure}
\label{sec:appendix_mpc}
Our low-level convex MPC controller is similar to previous works \cite{mitcheetahmpc, fast_and_efficient}, which includes a contact scheduler to determine the contact state of each foot (\emph{swing} or \emph{stance}), and separate controllers for swing and stance legs.
In addition, we implement a slope detector to react to changes in ground level, and an impedance controller to prevent excessive foot slipping.

\subsection{Phase-based Contact Scheduler}
The contact scheduler determines the contact state of each leg (swing or stance).
Similar to previous work by \citet{fast_and_efficient}, we adopt a phase-based representation for contact schedule. More specifically, we introduce four phase variables $\phi_{FR, FL, RR, RL}\in [0, 2\pi)$, one for each leg.
The subscripts denote the identity of each leg (Front-Right, Front-Left, Rear-Right, Rear-Left).
The phase of each leg denotes its progress in the current locomotion cycle.
As a leg moves, its phase $\phi$ increases monotonically from $0$ to $2\pi$, and wraps back to 0 as the next locomotion cycle starts.

At each control step, we determine the leg phases using the following procedure. First, we progress the phase of the front-right leg based on the stepping frequency (SF), $f$, which is obtained from the gait selector (Section~\ref{sec:gait_selector}):
\begin{align}
\phi_{FL}\gets \phi_{FL}+2\pi f \Delta t 
\end{align}
where $\Delta t$ is the control timestep (0.0025s).
We then determine the remaining leg phases according to the trotting pattern, where diagonal legs move together, and $180^{\circ}$ out-of-phase with the other diagonal:
\begin{align}
    \phi_{FL}&=\phi_{FR}+\pi \\
    \nonumber\phi_{RR}&=\phi_{FR}+\pi \\
    \nonumber\phi_{RL}&=\phi_{FR}
\end{align}

Depending on the phase of each leg, its motor command is computed by either the swing controller ($\phi<\pi$) or the stance controller ($\phi\geq \pi$).

\subsection{Swing and Stance Controller}
\label{sec:appendix_swing_stance}
We use separate control strategies for swing and stance legs, where the swing controller uses a PD control to track a desired swing foot trajectory, and the stance controller solves a model predictive control (MPC) problem to optimize for foot forces.
Our controller design is based on the work by \citet{mitcheetahmpc} with modifications. 
We briefly summarize them here. 
Please refer to the original work \cite{mitcheetahmpc} for further details.

\paragraph{Swing Controller} The goal of the swing controller is for swing legs to track a desired swing trajectory.
At each control step, the swing controller computes the swing trajectory by interpolating between three key positions, namely, the lift-off position $\vec{p}_{\text{lift-off}}$, the highest position in the air $\vec{p}_{\text{air}}$, and the landing position $\vec{p}_{\text{land}}$.
$\vec{p}_{\text{lift-off}}$ is the recorded lift-off position at the beginning of the swing phase. 
$\vec{p}_{\text{air}}$ is the foot's highest position in the swing phase, where the height is determined by the swing height (SH) from the gait selector (Section~\ref{sec:gait_selector}).
$\vec{p}_{\text{land}}$ is the estimated landing position computed by the Raibert Heuristic.
Once the desired swing trajectory is computed, the controller computes the leg's desired position based on its progress in the current swing state.
The controller then converts this desired position into joint angles using inverse kinematics, and uses a joint PD controller to track the desired position. 

\paragraph{Stance Controller} To find desired foot forces, the stance controller solves a MPC problem, where the objective is for the base to track a desired trajectory.
The trajectory is generated by numerically integrating the desired forward speed and steering speed, where the forward speed is provided by the speed policy (Section~\ref{sec:speed_policy}) and the steering speed is provided by an external teleoperator.
In addition, the height of the robot in this trajectory is determined by the base height (BH), which is provided by the gait selector (Section~\ref{sec:gait_selector}).
To solve this MPC problem efficiently, we cast it as a quadratic program (QP), where the objective is a quadratic trajectory tracking loss, and the constraints include the linearized robot dynamics, actuator limits and approximated friction cones.
Once the foot force is solved, we convert it into joint torque commands using jacobian transpose.
Note that the final joint torque command is the sum of this torque computed by MPC and an additional torque from position feedback.
Please see Appendix~\ref{sec:appendix_impedance_control} for details.

\subsection{Uneven Terrain Detection and Adaptation}
To operate in complex offroad terrains, it is crucial for the robot to adapt its base pose and foot swing range based on terrain shape.
Therefore, we implement an uneven terrain detector that estimates terrain orientation from foot contact information, and use it to adjust the robot's pose on steep slopes. 

\paragraph{Ground Orientation Estimation}
The orientation of the ground plane is estimated based on foot contact position and imu readings, similar to previous work by \citet{ethslopes}.
To find the ground orientation, we first find the ground normal vector in the robot frame, $\vec{n}_{\text{robot}}$, from foot contact positions.
More specifically, we keep track of the last contact position of each leg, $\vec{p}_{1,\dots,4}$, which is represented in \emph{robot frame}.
Assuming that the contact positions lie in the same ground plane, the relationship between the contact positions $\vec{p}_{1,\dots,4}$ and the ground normal vector $\vec{n}_{\text{robot}}$ can be represented by:
\begin{equation}
    \vec{n}_{\text{robot}}\tran\vec{p}_1=\vec{n}_{\text{robot}}\tran\vec{p}_2=\vec{n}_{\text{robot}}\tran\vec{p}_3=\vec{n}_{\text{robot}}\tran\vec{p}_4
\end{equation}

We find $\vec{n}_{\text{robot}}$ using the following least-squares formulation:
\begin{equation}
    \min_{\vec{n}_{\text{robot}}} \left\|\left[
        \begin{array}{c c c c}
             \vert & \vert & \vert & \vert\\
             \vec{p}_1 & \vec{p}_2 & \vec{p}_3 & \vec{p}_4\\
             \vert & \vert & \vert & \vert
        \end{array}
    \right]\tran 
    \vec{n}_{\text{robot}} - \vec{1}
    \right\|_2^2
\end{equation}

Finally, we convert the ground normal vector to the world frame based on the robot's orientation, which is provided by the onboard IMU.

\begin{figure}[t]
    \centering
    \begin{subfigure}[b]{0.4\linewidth}
        \centering
        \includegraphics[width=\linewidth]{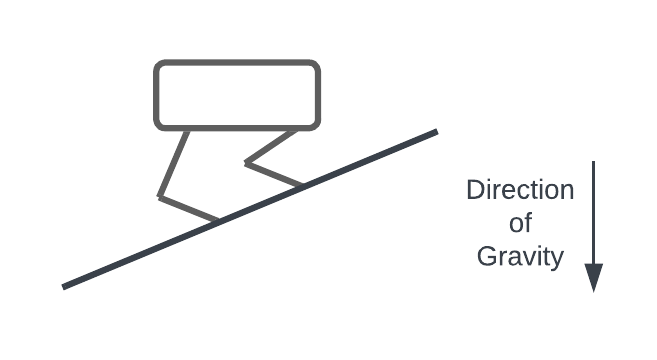}
        \caption{Water Level}
        \label{fig:stance_leg_water_level}
    \end{subfigure}
    \begin{subfigure}[b]{0.4\linewidth}
        \centering
        \includegraphics[width=\linewidth]{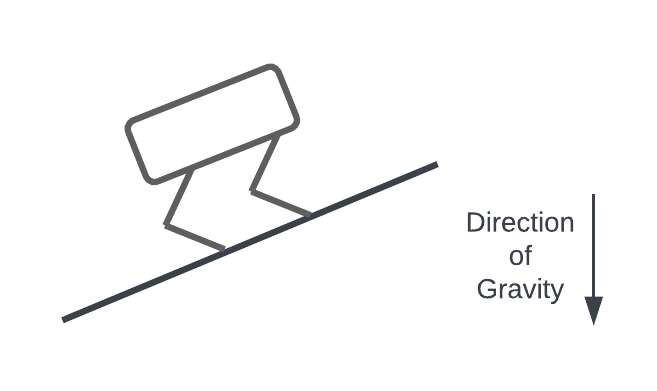}
        \caption{Ground Level}
        \label{fig:stance_leg_ground_level}
    \end{subfigure}
    \caption{\label{fig:slope_adaptation_stance_legs}Comparison of stance leg strategies on an upward slope. \textbf{Left}: If the robot remains water-level, its front legs are overly retracted, and its rear legs are overly extended, which makes it to control both legs. \textbf{Right}: If the robot remains ground-level, both front and rear legs have equal amount of extension. We keep the robot ground-level and convert the direction of gravity to ground frame based on estimated ground orientation.}
\end{figure}

\paragraph{Adaptation for Stance Legs}
To walk on uneven terrains, the robot needs to maintain its body to be \emph{ground-level}, instead of \emph{water-level}, so that each leg has equal amount of swing space (Fig.~\ref{fig:slope_adaptation_stance_legs}).
To achieve that, we express the base pose in the ground frame in the stance-leg MPC controller (Appendix~\ref{sec:appendix_swing_stance}), and adapt the direction of gravity based on the estimated ground orientation.

\paragraph{Adaptation for Swing Legs}                               
For swing legs, we adapt the \emph{neutral swing position} to be aligned with the direction of gravity. This configuration allows the base to be controlled more easily even on slippery surfaces, as shown in previous work by \citet{ethslopes}.

\subsection{Impedance Controller to Counter Foot Slipping}
\label{sec:appendix_impedance_control}
Since the MPC-based stance leg controller assumes static foot contacts, additional slip-handling technique is usually required for the robot to walk on slippery surfaces. 
For example, to prevent foot slipping, \citet{ethimpedance} implemented a probabilistic slip detector and used impedance control with different gains for slip and non-slip legs.
Similar to this work, we also implement an impedance controller to handle foot slips.
However, our approach adopts a unified gain for both slip and non-slip legs, and does not require a slip detector.

In our impedance controller design, the desired torque for each actuated DoF is the sum of the torque computed by MPC and an additional position feedback:
\begin{equation}
    \tau=\tau_{\text{MPC}} + k_p (q_{\text{ref}}-q)
\end{equation}
where $\tau_{\text{MPC}}$ is the desired torque computed by the MPC solver (Appendix~\ref{sec:appendix_swing_stance}), $q$ is the joint position, and $k_p$ is the position gain.
We set the reference, $q_{\text{ref}}$ to be the expected joint position as if the foot is in static contact with the ground.
Intuitively, when the foot is not slipping ($q\approx q_{\text{ref}}$), the additional torque from the position feedback is small, and our controller falls back to the standard MPC controller.
When the foot is slipping, the additional torque from position feedback will bring the leg to the non-slipping position. 
To determine $q_{\text{ref}}$, we first compute the desired position of the corresponding leg in Cartesian coordinates, $\vec{p}_{\text{ref}}$.
Assuming that the base is moving at the commanded velocity $\vec{\bar{v}}$ with no foot slip, the foot velocity in robot frame is just the negated base velocity, and the foot position can be computed by numerical integration:
\begin{align}   
    \vec{p}_{\text{ref}} &\gets \vec{p}_{\text{ref}}-\vec{\bar{v}}\Delta t
\end{align}
where $\Delta t$ is the control timestep. 
We then convert from foot position $\vec{p}_{\text{ref}}$ to joint position $q_{\text{ref}}$ using inverse kinematics.
\section{Additional Experiment Results}
\label{sec:appendix_experiment}
\subsection{Generalization to Unseen Terrain Instances}
\label{sec:appendix_generalization}
\begin{table}[t]
    \centering
    \small
    \begin{tabular}{c|c c c c}
    \hline
         & Trail 1 & Trail 2 & Trail 3 & Trail 4 \\\hline
        Trail Length (km) &0.45&0.41&0.2&0.51\\\hline
        Terrain Type &Dirt&Mixed&Asphalt& Mud\\\hline
        Average Speed (m/s) &0.59&0.74&0.94&0.59\\\hline
        
    \end{tabular}
    \caption{Summary of test trails. Our framework selects different locomotion skills (speed and gait) based on terrain type.}
    \label{tab:long_range}
\end{table}

\begin{figure}[t]
    \centering
    \vspace{-2em}
    \includegraphics[width=0.8\linewidth]{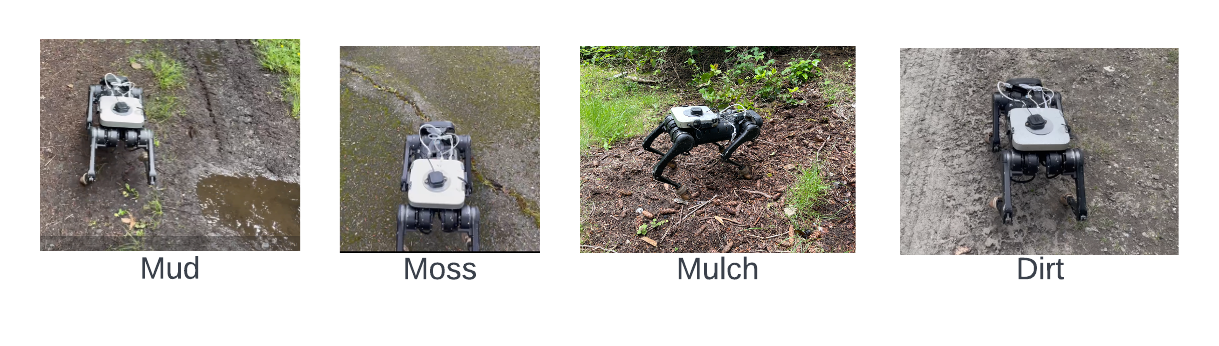}
    \vspace{-1.5em}
    \caption{Our framework generalizes to unseen terrain types, such as mud, moss, mulch and dirt.}
    \label{fig:unseen_terrains}
\end{figure}
\begin{figure}[b]
    \centering
    \includegraphics[width=0.9\linewidth]{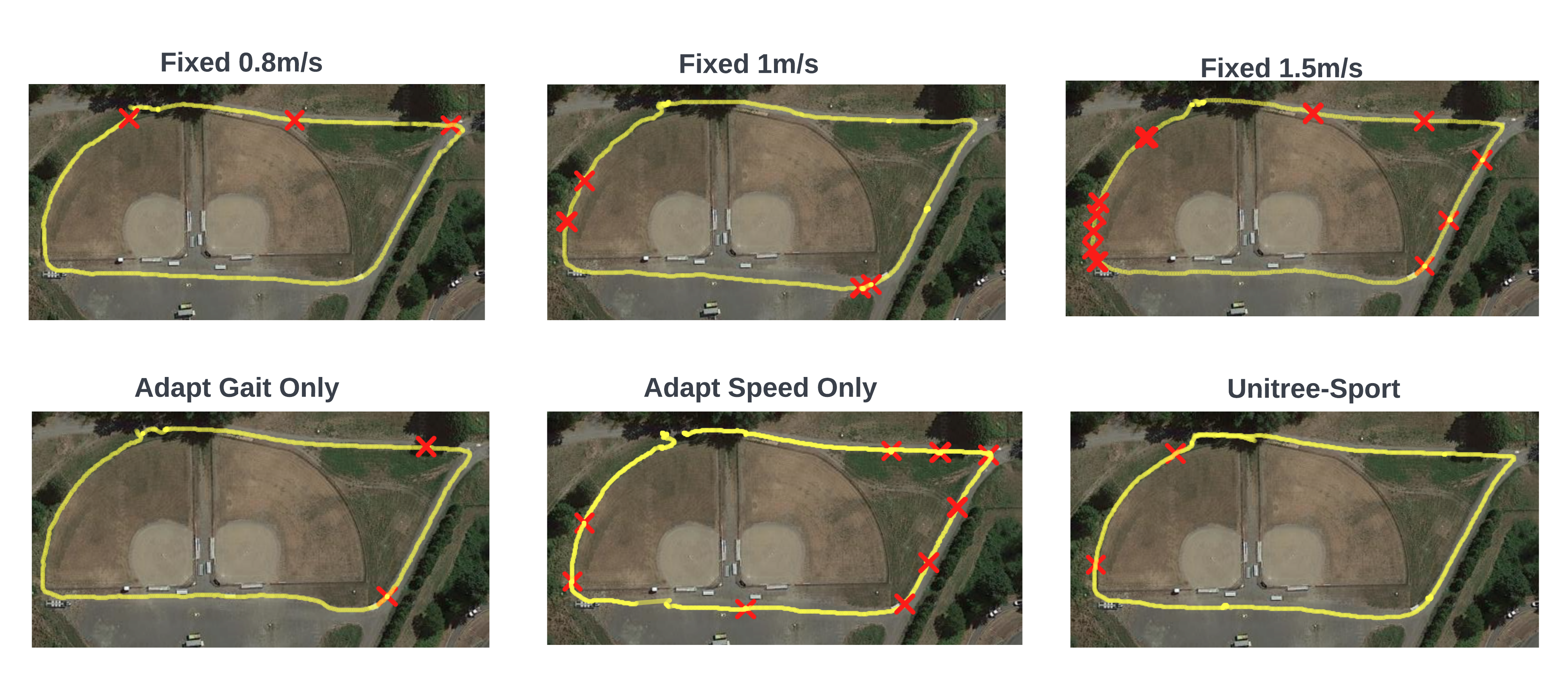}
    \caption{GPS logs (yellow) and failure locations (red cross) for the policies tested in Table.~\ref{tab:ablation_table}}
    \label{fig:failure_locations}
\end{figure}

We test the performance of our framework in a number of trails not seen during training. 
Please see Table~\ref{tab:long_range} for some examples.
These trails include a number of terrain types that are not seen during training, such as mud, moss, mulch and dirt (Fig.~\ref{fig:unseen_terrains}).
Our framework generalizes well to these terrains, and enables the robot to traverse through them quickly and safely.

\subsection{Details about Robot Failures}
\label{section:failure_locations}

For each policy tested in the ablation study, we plot its GPS tracking and failure locations in Fig.~\ref{fig:failure_locations}.
We find that the incorrect selection of speed and gait can both lead to robot failures.
For example, a faster forward speed makes the robot more susceptible to small unevennesses on the ground. Therefore, non-adaptive policies with higher speeds fail more frequently, especially on areas such as grass or gravel.
Moreover, gait parameters such as swing frequency and swing height also affects the controller's stability. For example, a low swing height (SH) can get the legs trapped in deep grass, and a low swing frequency (SF) can result in large amount of foot swings at each step, which can go beyond the robot's capability.

\subsection{Ablation Study on Perception Module}
\label{sec:appendix_perception}
\paragraph{Model Architecture and Training}

We use the FCHarDNet-70 architecture \cite{fchardnet}, which is obtained from the paper's open-sourced code-base. 
For pre-training on RUGD dataset \cite{RUGD}, we train the model for 100 epoches with a batch size of 10 using the Adam Optimizer where the learning rate is set to 0.001.
For more robust training, we augment the images from RUGD with random crops and color adjustments.
For fine-tuning on demonstration data, we train the model for 60 epoches with a batch size of 32, using the Adam Optimizer with the same learning rate of 0.001.
Both the pre-training and fine-tuning are conducted on a desktop computer with a Nvidia 2080Ti GPU, where pre-training takes around 6 hours and fine-tuning takes around 20 minutes.

\paragraph{Baselines}
As discussed in Section.~\ref{sec:semantic_embedding}, we train the speed policy by finetuning on the pixel-wise semantic embedding, which is extracted from the output of the last hidden layer. To justify this design choice, we compare our way of training the speed policy with two baselines. For the model \emph{finetuned on class labels}, we extract the embedding of each pixel from the one-hot encoding of the model's predicted semantic class. For the model \emph{trained from scratch}, we train the FCHarDNet from scratch on the demonstration data without pre-training.

\paragraph{Results}

\begin{table}[t]
    \small
    \centering
    \begin{tabular}{c|c|c|c}
        \hline
          & Finetuned on Hidden Layers& Finetuned on Class Label & Trained from Scratch \\\hline
        Validation Loss & $\mathbf{0.061\pm 0.002}$ & $0.075\pm 0.003$ & $0.088\pm 0.013$\\\hline
    \end{tabular}
    \caption{Comparison of performance on different ways of training the speed policy.}
    \label{tab:perception_generalization}
\end{table}

\begin{figure}[t]
    \centering
    \includegraphics[width=0.9\linewidth]{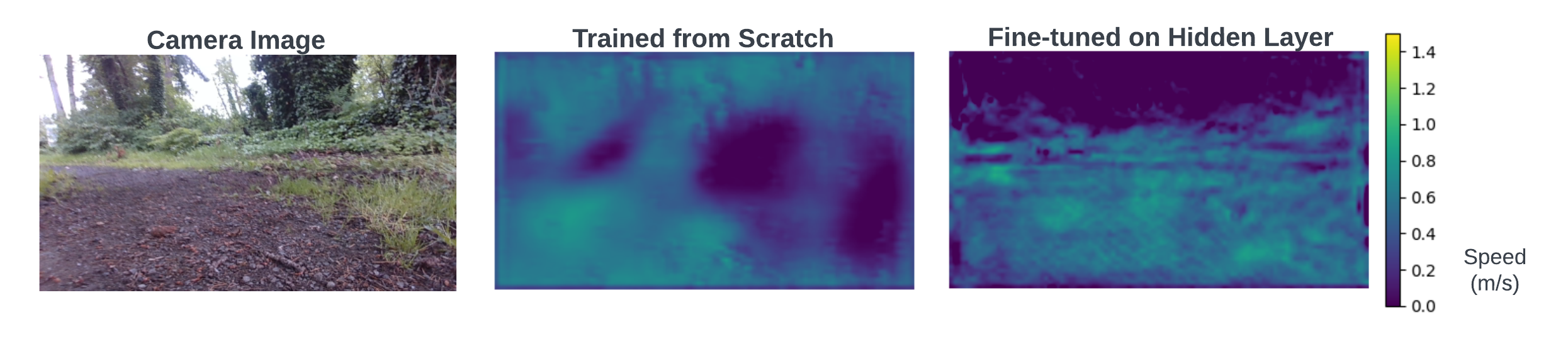}
    \caption{Comparison of different ways to predict the speed map. \textbf{Left}: the camera image contains multiple semantic classes including mulch, grass and trees. \textbf{Middle}: the speed model trained from scratch has a low resolution and cannot identify different semantic classes. \textbf{Right}: the speed model fine-tuned from semantic embedding accurately identifies different terrain types and computes the desired speed for each terrain.}
    \label{fig:perception_comparison}
\end{figure}
We train our method and the baselines on the demonstration data, and test the model's performance on a small validation set, where the data is collected on a different trail. For each model, we repeat the experiment 5 times with different random seeds and report the mean and standard deviation of the loss function (mean-squared loss).
The result is summarized in Table.~\ref{tab:perception_generalization}. 
Our method, which finetunes on the output from hidden layer, achieves the lowest error on the validation set.
The model fine-tuned on class label achieves a big loss on both datasets.
This is likely due to noisy label prediction, which results from the distribution shift between the model's training data (RUGD) and testing data (robot images).
Since the model trained from scratch tunes the entire FCHarDNet on the small set of demonstration data, it overfits to the training data and does not generalize well to the validation set.
Moreover, a closer look at the models' predictions shows that the model trained from scratch predicts a blurry speed map with incorrect speed predictions for several regions, compared to our fine-tuned model (Fig.~\ref{fig:perception_comparison}). 
This is likely due to the lack of granularity in demonstration data, which only labels the desired average speed over a fixed region.

\end{document}